\documentclass{article}
\usepackage{spconf,amsmath,graphicx}
\usepackage{todonotes,comment}
\usepackage{amsmath,amssymb,amsfonts}
\usepackage{listings}
\usepackage[hyphens]{url}

\title{Mango: A Python Library for Parallel Hyperparameter Tuning}

\name{Sandeep Singh Sandha$^{\star}$ 
\quad Mohit Aggarwal$^{\dagger}$ 
\quad Igor Fedorov$^{\dagger}$ 
\quad Mani Srivastava$^{\star}$
\thanks{The research reported in this paper was supported by a summer internship at Arm. The first and the last authors would also like to acknowledge the support of their research from the CONIX Research Center, one of six centers in JUMP, a Semiconductor Research Corporation (SRC) program sponsored by DARPA. The views and conclusions contained in this document are those of the authors and should not be interpreted as representing the official policies, either expressed or implied, of the funding agencies.
}
\thanks{© 20XX IEEE.  Personal use of this material is permitted.  Permission from IEEE must be obtained for all other uses, in any current or future media, including reprinting/republishing this material for advertising or promotional purposes, creating new collective works, for resale or redistribution to servers or lists, or reuse of any copyrighted component of this work in other works.}
}
\address{$^{\star}$ University of California, Los Angeles
			\\ $^{\dagger}$Arm Research
			}
			    
\begin{document}
%
\maketitle

\begin{abstract}
Tuning hyperparameters for machine learning algorithms is a tedious task, one that is typically done manually. To enable automated hyperparameter tuning, recent works have started to use techniques based on Bayesian optimization. However, to practically enable automated tuning for large scale machine learning training pipelines, significant gaps remain in existing libraries, including lack of abstractions, fault tolerance, and flexibility to support scheduling on any distributed computing framework. To address these challenges, we present \textit{Mango}, a Python library for parallel hyperparameter tuning.  Mango enables the use of any distributed scheduling framework, implements intelligent parallel search strategies, and provides rich abstractions for defining complex hyperparameter search spaces that are compatible with scikit-learn. Mango is comparable in performance to Hyperopt~\cite{bergstra2013hyperopt}, another widely used library. Mango is available open-source~\cite{Mango} and is currently used in production at Arm Research to provide state-of-art hyperparameter tuning capabilities.

\end{abstract}


\section{Introduction}
\label{sec:intro}
Although machine learning (ML) classifiers have enabled a variety of practical use cases, they are highly sensitive to the choice of hyperparameters~\cite{bergstra2013hyperopt,feurer2015efficient}. Many classifiers are characterized by complex and large hyperparameter spaces. For example, a random forest classifier is parameterized by the number of trees and their depth, the number of features per tree, and the minimum samples per leaf node, to name a few. Hyperparameter tuning is an outer optimization loop on top of ML model training. Therefore, evaluating a given hyperparameter configuration is costly, ranging from hours to days for large data sets~\cite{bergstra2013hyperopt}. In this setting, traditional brute force techniques like the random, grid, and sequential search are not efficient and may require a significant amount of time to find the optimal hyperparameters~\cite{desautels2014parallelizing, groves2018efficient,thornton2013auto,bergstra2012random}. Due to the increasing use of automated ML pipelines, which involve training models at large scale on distributed computing clusters without a developer in the loop, there is a need for efficient and horizontally scalable hyperparameter optimization frameworks. 

Sequential Model Bayesian Optimization (SMBO) is an efficient method for hyperparameter optimization due to its ability to find optima with fewer evaluations compared to the grid or random search~\cite{feurer2015efficient}. While SMBO is effective, it can be suboptimal in a distributed computing environment, necessitating research into parallel Bayesian optimization algorithms~\cite{desautels2014parallelizing, groves2018efficient}. In practice, however, there does not exist a software library that implements these algorithms efficiently, provides easy-to-use abstractions for defining hyperparameter search spaces, and can take advantage of any distributed computing framework for scheduling configuration evaluations. 

To enable efficient parallel hyperparameter search, we present \textit{Mango}. Mango is an open source~\cite{Mango} Python library designed for ease of use and extensibility. Internally, Mango implements a state of the art optimization algorithm based on batch Gaussian process bandit search using upper confidence bound as the acquisition function~\cite{desautels2014parallelizing}. Mango implements adaptive exploitation vs. exploitation trade-off as a function of search space size, number of evaluations, and parallel batch size. Mango has the following novel features that are not available in existing hyperparameter optimization libraries:

\begin{itemize}
  \item Intelligent parallel search strategy based on Batched Gaussian Process Bandits~\cite{desautels2014parallelizing} and clustering~\cite{groves2018efficient}, such that information gain is maximized.
  \item Modular design that decouples the optimizer from the scheduler to allow the use of any distributed task scheduling framework. This is especially important for integration with ML pipelines running on distributed clusters.
  \item Rich abstraction to easily define complex hyperparameter search spaces that are compatible with scikit-learn's model\_selection functions, and scipy.stats distributions. 
\end{itemize}

This paper gives an overview of the Mango library, the design choices, and architecture. We discuss the implemented parallel optimization algorithms, along with the realization of Mango on a Kubernetes cluster using Celery. Mango is designed as a Black-Box optimizer that can be used to optimize any function. At present, our development is tailored towards tuning the hyperparameters of traditional ML classifiers. Mango is deployed in production at Arm Research to provide hyperparameter tuning for ML jobs in a cluster setting. For evaluation, we compare Mango with Hyperopt~\cite{bergstra2013hyperopt} and show that the optimization algorithms of Mango are comparable in performance. 

\textbf{Related work}:
There are several existing libraries designed for hyperparameter tuning which come under the umbrella of automatic ML (AutoML) systems, including Hyperopt~\cite{bergstra2013hyperopt}, Auto-sklearn~\cite{feurer2015efficient} and Auto-WEKA~\cite{thornton2013auto}. These libraries provide limited support for parallel search using distributed scheduling frameworks. For example, the parallel execution in Hyperopt~\cite{bergstra2013hyperopt} is limited due to its dependence on the MongoWorker~\cite{hyperopt_1} processes and Apache Spark. The parallel execution in Auto-sklearn~\cite{feurer2015efficient} requires users to manually start multiple instances on a shared file system~\cite{auto_sklearn_1}. Auto-WEKA~\cite{thornton2013auto} runs multiple sequential algorithms with different random seeds for parallel execution on a single compute node.  
The engineering effort required to integrate these libraries into existing ML training pipelines deployed at Arm would be prohibitive due to their lack of abstraction and inability to support arbitrary distributed task scheduling frameworks. As a result, Mango was developed in-house with a modular design for ease of integration with any distributed task scheduling framework.

\vspace{-0.04in}
\begin{lstlisting}[language=Python,caption={Sample hyperparameter search space definition for XGBoost's XGBClassifier},captionpos=b, label={lst:space_xgboost}]
"learning_rate":   uniform(0, 1),
"gamma":           uniform(0, 5),
"max_depth":       range(1,10),
"n_estimators":    range(1,300),
"booster":['gbtree','gblinear','dart']
\end{lstlisting}

\textbf{Hyperparameter Tuning}: To illustrate the complexity of the hyperparameter tuning problem, consider tuning XGBoost’s~\cite{XGBoost}  XGBClassifier. 
Considering only the five hyperparameters shown in listing~\ref{lst:space_xgboost}, the cardinality of the search space is on the order of $10^6$. As such, manual and grid search become prohibitively expensive.
Moreover, Section~\ref{sec:evaluation} will show that random search does not yield favorable results on this large search space. To automatically explore such large search spaces with the minimum number of evaluations, researchers have proposed intelligent search strategies based on Bayesian optimization. Mango takes advantage of such approaches to traverse the search space in a distributed computing environment, all the while offering easy-to-use abstraction for defining parameter space.   

\section{Mango Architecture}
\label{sec:mango}
\begin{figure}
    \centering
    \includegraphics[width=\linewidth]{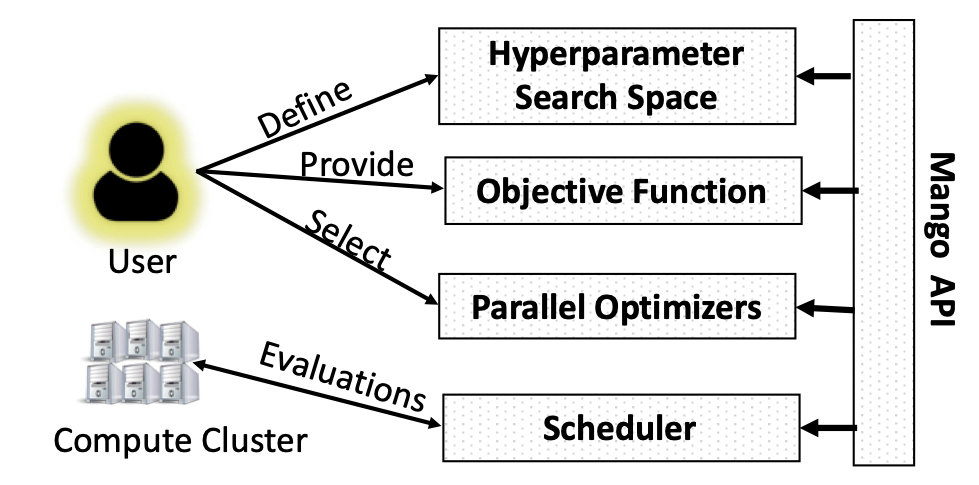} 
    \caption{Mango Architecture and User Workflow: (1) User defines hyperparameter search space, (2) User provides the desired objective function, (3) User selects the parallel optimization algorithm, and (4) Mango selects the default scheduler or uses user provided implementation that evaluates the objective function.
     }
      \vspace{-0.04in}
    \label{fig:mango_arch}
\end{figure}

The architecture and workflow of Mango are shown in Fig.~\ref{fig:mango_arch}. 
The usage workflow is enabled by four abstractions provided by the Mango API. Next, we discuss the details of these abstractions with their example usage.

\vspace{-0.04in}
 \begin{lstlisting}[language=Python,caption={Hyperparameter search space in Mango for SVM.},captionpos=b, label={lst:space_svm}]
from scipy.stats import uniform
from mango.domain.distribution 
            import loguniform
param_dict={"gamma":uniform(0.1, 4),
              "C":loguniform(-7, 7)}
\end{lstlisting}
\vspace{-0.04in}
\subsection{Hyperparameter Search Space}
\label{ssec:searchspace}
The hyperparameter search space defines the domain of the search problem.  Hyperparameters can be both continuous-valued or discrete/categorical and have different distributions.
Mango provides easy-to-use domain space definitions, which are also compatible with the scikit-learn~\cite{scikit-learn} model selection functions. The hyperparameter search space is defined using a Python dictionary with parameter names (string) as keys mapping to the associated parameter range. The search space range can either use distributions or Python constructs (list and range). We support all (70+) distributions from scipy.stats~\cite{scipy} and also provide the capability to define new distributions. Distributions must provide a method for sampling from the distribution (such as those from scipy.stats.distributions). New distributions can be defined by extending the scipy distribution constructs. We also provide a predefined, widely used \textit{loguniform} distribution as an example. The search choices within lists and ranges are sampled uniformly. Categorical and discrete hyperparameters are defined using a list of strings.

Listing~\ref{lst:space_svm} shows the sample hyperparameter space definition for a support vector machine (SVM) classifier, with two hyper parameters: \textit{C} and \textit{gamma} using \textit{uniform} distribution from scipy and  \textit{loguniform} distribution from Mango. Hyperparameter space definitions can be arbitrarily large and complex, as shown in listing~\ref{lst:space_xgboost}.

\subsection{Objective Function}
\label{ssec:objective}
Mango's goal is to find the optimal value for the specified objective function within the bounds of the search space defined by hyperparameter definitions. Mango allows the flexibility to define arbitrary objective functions. The Mango optimizer selects a batch of configurations for evaluation, which is passed as a list argument to the objective function. The example in listing~\ref{lst:objective_serial} shows a sample skeleton of an objective function which serially evaluates the list of hyperparameters and returns the list of the successful objective function values and their respective hyperparameters. 
\vspace{-0.04in}
\begin{lstlisting}[language=Python,caption={Skeleton of the serial objective function in Mango.},captionpos=b,label={lst:objective_serial}]
def objective_function(params_list):
    evals = []
    params = []
    for par in params_list:
        eval = Schedule_Classifier(par)
        evals.append(eval)
        params.append(par)
    return evals, params
\end{lstlisting}

Listing~\ref{lst:objective_serial} represents the most common usage in the single node setting, where the objective function evaluations are done sequentially. Since the objective function has access to a batch of configurations, external distributed frameworks can be used to evaluate the batch in parallel. To use all cores in local machine, threading can be used to evaluate a set of values. More details on the scheduler are discussed in Section~\ref{ssec:scheduler}. When defining the objective function, the user also has the flexibility to record any intermediate debugging results by saving them to the disk or storing them in global variables. The choice to have the objective function consume batches of configurations and be independent of the choice of scheduler was motivated by the desire to decouple scheduler and optimizer abstractions.

\subsection{Parallel Optimization Algorithms}
\label{ssec:algorithms}

The optimization algorithms in Mango are based on widely used Bayesian optimization techniques, extended to sample a batch of configurations in parallel. Currently, Mango provides two parallel optimization algorithms that use the upper confidence bound as the acquisition function. The first algorithm is motivated by the research of Desautels et al.~\cite{desautels2014parallelizing}. In the second algorithm, we create clusters of acquisition function in spatially distinct search spaces and select the maximum value within each cluster to create the batch. The clustering approach is based on ~\cite{groves2018efficient}. We do the necessary modifications to have an efficient, practical implementation for both of these algorithms.

To maximize the acquisition function, we use the Monte Carlo method by randomly sampling the search space using the hyperparameter distributions and then finding the maximum value from these samples. Mango internally selects the number of random samples using a heuristic based on the number of hyperparameters, search space bounds, and the complexity of the search space itself. This also ensures that the acquisition function is evaluated at valid configurations only. This is especially useful for proper treatment of discrete and categorical variables as described in ~\cite{2018arXiv180503463G}, although they modify the kernel function to get the same behavior. In addition to the implemented parallel algorithms, Mango also supports a random optimizer which selects a batch of random configurations.

\subsection{Scheduler}
\label{ssec:scheduler}
The scheduler is used to evaluate batches of configurations. The scheduler can process the entire batch in parallel or can control the level of parallelism depending on available resources. The maximum level of parallelism per job is decided by the size of the batch, which is a user-controlled parameter. For distributed scheduling frameworks to handle the case of straggler/faulty workers, only a partial set of evaluated hyperparameter values can be returned by the objective function, as shown in listing~\ref{lst:objective_serial}. The objective function returns both the list of objective function values (\textit{evals}) and the respective configuration (\textit{params}), so as to account for out of order evaluation and missing evaluations, which often happen in large scale deployments. 

\vspace{-0.04in}
\begin{lstlisting}[language=Python,caption={Skeleton of the Celery based parallel objective function in Mango.},captionpos=b,label={lst:objective_parallel}]
def objective_celery(params_list):
    process_queue = []
    for par in params_list:
        process = train_clf.delay(par)
        process_queue.append((process, par))
    evals = []
    params = []
    for process, par in process_queue:
        result = process.get()
        evals.append(result)
        params.append(par)
    return evals, params
\end{lstlisting}

Mango is not tied to any scheduling framework. This choice was left to the user and deployment setting. We provide several examples in~\cite{Mango_examples}, some of which use local schedulers and others a distributed scheduler. The \textit{schedule\_classifier} call in listing~\ref{lst:objective_serial} represents traditional serial execution. For local usage, users can directly compute the classifier accuracy in the objective function definition itself as shown in the SVM example in \textit{SVM\_Example.ipynb}~\cite{Mango_examples}, 
whereas external frameworks can be used in a distributed cluster setting. An example of using Celery, a distributed task queue, is shown in listing~\ref{lst:objective_parallel}.
A simple starting point is available in the \textit{KNN\_Celery.ipynb}~\cite{Mango_examples}
which optimizes a k-nearest neighbors classifier. Mango is deployed in a production system at Arm Research using Celery~\cite{celery} deployed on a Kubernetes cluster.

There are several user-controlled options in Mango. For example, the batch size, the choice of algorithm (hallucination, clustering, and random), the maximum number of allowed iterations, and the number of initial random evaluations. The heuristic-based search space size used to maximize the acquisition function can also be overridden by the user. 
More details are in Mango's documentation~\cite{Mango}.

\begin{figure}
    \centering
    \includegraphics[width=\linewidth]{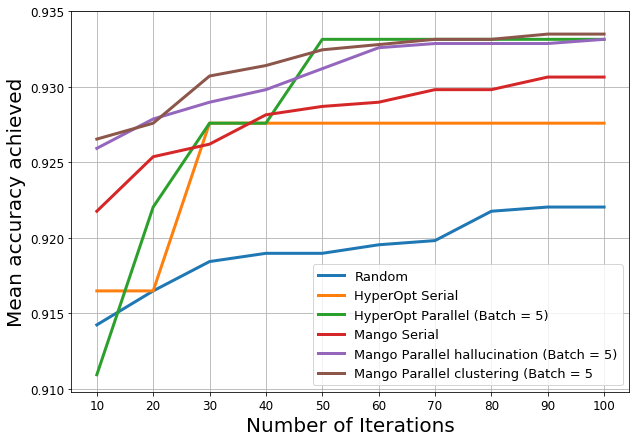} 
    \caption{Comparison of optimization algorithms of Mango with Hyperopt and random strategy for XGBClassifier classifier. Results averaged over 20 experiments. Number of iterations denotes the number of batches evaluated. For serial experiments, batch size is 1.}
    \label{fig:mango_evaluation_1}
\end{figure}

\section{Evaluation}
\label{sec:evaluation}
In this section, we compare the performance of algorithms implemented in Mango with the widely used library Hyperopt. Our goal is to show that Mango offers optimization performance comparable to state of the art libraries like Hyperopt, with the additional benefit of easy adoption for parallel hyperparameter tuning pipelines. First, we evaluate tuning the hyperparameters of XGBoost's XGBClassifier model on the wine dataset~\cite{Dua}. The parameter search space is given in listing~\ref{lst:space_xgboost}.

\begin{figure}
    \centering
    \includegraphics[width=\linewidth]{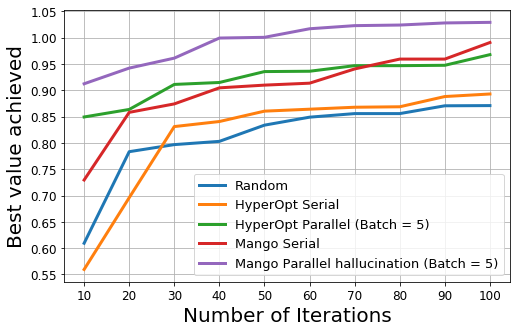} 
    \caption{Comparison of optimization algorithms of Mango with Hyperopt and random strategy for modified Branin function. Results averaged over 10 experiments.}
    \label{fig:mango_evaluation_2}
\end{figure}

The evaluation result is presented in Fig.~\ref{fig:mango_evaluation_1}. 
All of the approaches perform better than random search. Results for serial Mango are obtained by setting batch size equal to 1. Comparing Hyperopt serial with Mango serial, the performance of Mango serial is slightly better. For parallel evaluation, we run the algorithms with batch size of 5. 
Both parallel algorithms implemented in Mango perform slightly better than the Hyperopt parallel algorithm, especially when the number of iterations is limited to $40$ or less. As the number of iterations increases, the performance of Mango and Hyperopt converges.

Next, we evaluate using the Branin function, which is a common benchmark for optimization algorithms~\cite{jones2001taxonomy}. We consider a modified Branin function with mixed discrete and continuous variables~\cite{halstrup2016black}. The parameter search space and the exact function definition used are available online in \textit{Branin\_Benchmark.ipynb}~\cite{Mango_examples}.
For this evaluation, we only run the hallucination based parallel algorithm in Mango. We repeat the experiment $10$ times and report average results in Fig.~\ref{fig:mango_evaluation_2}. In both the serial and parallel regimes, Mango outperforms Hyperopt.
Several other examples using Mango to tune classifier hyperparameters are available online~\cite{Mango_examples}.

\section{Conclusion}
\label{sec:Conclusion}

In this paper, we presented Mango, 
a parallel hyperparameter tuning library designed for modern ML training pipelines deployed on distributed computing clusters.
We presented the abstraction available in Mango, which can easily define complex hyperparameter search spaces, the desired objective function, and the optimization algorithm. With Mango, the user is not tied to a specific scheduling framework and can evaluate the batch of configurations using any platform of choice. 
Experimental results show Mango achieves state of the art optimization results on two standard tasks. Mango is available as an open-source Python library~\cite{Mango}, is deployed in production at Arm Research, and is continuously being tested using new classifiers and datasets. Our future goal is to add more parallel optimization algorithms.


\bibliographystyle{IEEEbib}
\bibliography{refs}

\end{document}